\pdfoutput=1
\documentclass{article}

\usepackage[final]{ewrl_2026}

\usepackage[utf8]{inputenc} %
\usepackage[T1]{fontenc}    %
\usepackage{hyperref}       %
\usepackage{url}            %
\usepackage{booktabs}       %
\usepackage{amsfonts}       %
\usepackage{nicefrac}       %
\usepackage{microtype}      %
\usepackage{xcolor}         %

\title{Towards Understanding the Impact of Plasticity Loss on Reinforcement Learning in Stochastic Environments}

\author{%
  Philipp Bordne \\
  University of Freiburg \\
  \texttt{bordnep@cs.uni-freiburg.de} \\
  \And
  André Biedenkapp\thanks{Work done while at the University of Freiburg.} \\
  Karlsruhe Institute of Technology \\
  \texttt{biedenkapp@kit.edu} \\
}

\usepackage[acronym]{glossaries}
\usepackage{subcaption}
\usepackage{amsthm}
\usepackage{mathrsfs}
\usepackage{wrapfig}

\usepackage{booktabs}
\usepackage{xcolor}
\usepackage{todonotes}

\usepackage{amsmath,amsfonts,bm}

\def\eqref#1{equation~\ref{#1}}

\def\1{\bm{1}}

\def\ra{{\textnormal{a}}}

\def\ro{{\textnormal{o}}}

\def\va{{\bm{a}}}
\def\vb{{\bm{b}}}

\def\vo{{\bm{o}}}

\def\vx{{\bm{x}}}
\def\vy{{\bm{y}}}

\def\mD{{\bm{D}}}

\DeclareMathAlphabet{\mathsfit}{\encodingdefault}{\sfdefault}{m}{sl}
\SetMathAlphabet{\mathsfit}{bold}{\encodingdefault}{\sfdefault}{bx}{n}

\def\gA{{\mathcal{A}}}

\def\gM{{\mathcal{M}}}
\def\gN{{\mathcal{N}}}
\def\gO{{\mathcal{O}}}
\def\gP{{\mathcal{P}}}

\def\gS{{\mathcal{S}}}

\def\gU{{\mathcal{U}}}

\def\sN{{\mathbb{N}}}

\def\sR{{\mathbb{R}}}

\newcommand{\E}{\mathbb{E}}

\usepackage{cleveref}
\newtheorem{definition}{Definition}

\newacronym{rl}{RL}{Reinforcement Learning}
\newacronym{mdp}{MDP}{Markov Decision Process}
\newacronym{pomdp}{POMDP}{Partially Observable Markov Decision Process}
\newacronym{nn}{NN}{neural network}
\newacronym{dmc}{DMC suite}{DeepMind Control suite}
\newacronym{ln}{LN}{layer normalization}
\newacronym{sac}{SAC}{Soft Actor-Critic}
\newacronym{a_noise}{$\sigma_\text{a}$}{action noise}
\newacronym{o_noise}{$\sigma_\text{o}$}{observation noise}
\newacronym{iqm}{IQM}{inter-quartile mean}
\newacronym{ci}{CI}{confidence interval}
\newacronym{td}{TD}{temporal difference}

\begin{document}

\maketitle
\setcounter{footnote}{0}

\begin{abstract}
  Deep reinforcement learning (RL) is a powerful framework for sequential decision-making. Despite many high-profile successes in recent years, deep RL is known to be brittle. An emerging line of research attributes this brittleness to a loss of plasticity in neural networks. However, plasticity loss has so far been understudied in stochastic environments. We study this interplay systematically, training SAC agents on three DeepMind Control tasks augmented with controlled levels of observation and action noise. We find that plasticity-preserving methods increase learning robustness on the two harder tasks, with their advantage growing as observation noise increases, while combining resets with layer normalization underperforms resets alone at every observation noise level tested. Further, stochasticity can accelerate the loss of plasticity in harder environments, whereas in easier settings noise can act as a regularizer that sustains plasticity. Lastly, plasticity loss is most pronounced in the early stages of training. In the most challenging stochastic settings, TD instability rather than gradual plasticity loss becomes the dominant failure mode.
\end{abstract}

\section{Introduction} \label{sec:introduction}

Over the last decade, deep \gls*{rl} \citep{sutton-mit18a} has achieved impressive results from game playing \citep{mnih-nature15a,grooten-aaai26a} over energy markets \citep{harder-energyai23a} to control of complex real world systems \citep{bellemare-nature20a,degrave-nature22a,kaufmann-nature23a}.
Predominantly however, RL research still is focused on simulated environments with deterministic transition dynamics \citep[][]{todorov-iros12a,brockman-arxiv16a,tassa-arxiv18a}.
However, stochasticity is very much inherent in real-world applications, where sensors are noisy, actuators imprecise, and dynamics uncertain.
A natural question is therefore how well modern deep \gls*{rl} algorithms learn under such conditions, and what the root cause is for the observed brittleness of \gls*{rl} \citep{henderson-aaai18a} in the presence of noise.

Multiple lines of research have emerged to understand and mitigate the brittleness of \gls*{rl}.
On the side of mitigation, many works address this by automating design choices present in \gls*{rl} pipelines \citep{parkerholder-jair22a}.
Some work in this direction is more concerned with augmenting environments, e.g., via curriculum learning \citep{openai-arxiv19a,klink-neurips20a} to expose agents to more diverse experiences, while others are trying to address the brittleness via hyperparameter tuning \citep{eimer-icml23a}.
While such works are able to produce more stable and performative \gls*{rl} agents, they provide little insight into the ultimate causes.

On the side of understanding the causes, attention has shifted towards the \textit{loss of plasticity} of neural networks in deep \gls*{rl} pipelines: agents are known to overfit to transitions collected early in training, when the policy is still far from optimal \citep{nikishin-icml22a,igl-iclr21a}, gradually losing the ability to update their representations in response to new experience \citep{lyle-icml23a}.
Interventions that counteract this, such as periodic resets and normalization, measurably improve learning on standard benchmarks \citep{nikishin-icml22a,nauman-icml24a}.
Closely related work has attributed the primacy bias to \gls*{td} instability, i.e. the early divergence of $Q$-values \citep{hussing-rlj24a}.

The interplay between plasticity loss and environment stochasticity remains largely unexamined, although two considerations suggest it warrants study.
The first is observational: in a benchmarking study of robust \gls*{rl} algorithms, \citet{glossop-neurips22a} find that the presence of noise \textit{during training} is considerably more detrimental than the same noise at deployment.
The second is mechanistic: \gls*{rl} algorithms optimize an \textit{expected} return, and the samples used to estimate it are collected throughout training.
Zero-mean noise should therefore average out, provided realizations are weighted equally regardless of when they were collected.
Plasticity loss might break this, inducing a bias towards early samples.
Accordingly, our object of study is \textit{learning} robustness, the ability of an algorithm to learn a task at all from noisy samples, rather than robustness in the more common sense of a deployed controller that remains stable or safe under disturbances \citep{brunke-arcras22a}.

This motivates our central question: \textit{can preserving plasticity improve learning robustness?}
We investigate this with a systematic study of \gls*{sac} \citep{haarnoja-icml18a} on three proprioceptive control tasks from the \gls*{dmc} \citep{tassa-arxiv18a}, augmented with controlled levels of observation and action noise.
We do not verify the mechanism directly, but our setting yields a testable prediction: if plasticity loss is more damaging under noise, the advantage of plasticity-preserving methods should grow with the noise level.
We probe plasticity throughout training to verify that the tested methods indeed preserve it and to establish plasticity as a potential factor in learning robustness under noise.
We additionally track maximum $Q$-values to separate the effects of plasticity loss and \gls*{td} instability \citep{hussing-rlj24a}.

Our key findings are:
\begin{itemize}
    \item Plasticity-preserving methods --- in particular periodic resets \citep{nikishin-icml22a} --- improve learning robustness under observation noise on the two harder tasks, with their advantage growing at higher noise levels.
    \item The relationship between noise and plasticity loss is bidirectional as noise accelerates plasticity loss, and low plasticity in turn compounds learning failure. On easier tasks, however, noise can act as a regularizer that actually sustains plasticity.
    \item Plasticity loss is most pronounced in the early stages of training and is exacerbated by noise, but in the most challenging settings the dominant failure mode is \gls*{td} instability rather than gradual plasticity loss and stochastic environments make this catastrophic failure mode more likely to occur.
\end{itemize}

\section{Related Work}
\paragraph{Controlled noise in simulated learning tasks.}
Previous work has augmented simulators with observation and action noise to reflect stochastic dynamics in real world applications, with the objective of obtaining policies that bridge the sim-to-real gap \citep{peng-icra18a} or to study the robustness of learned policies under disturbances \citep{yuan-ral22a}.
\citet{farebrother-icml24a} compare categorical and MSE regression on the Arcade-Learning-Environment \citep{machado-jair18a} with (default) and without action-noise.
They find that categorical regressions' improved performance over MSE is contingent upon the presence of action noise and attribute it to being less prone to overfitting on noisy targets.
\citet{glossop-neurips22a} study how different noise types (dynamics, observation, action) affect the score achievable by PPO \citep{schulman-arxiv17a} and SAC \citep{haarnoja-icml18a} on the cart-pole environment and find noise on dynamics and observations to be much more detrimental to performance. Interestingly, they find that the presence of observation noise during training is already detrimental to performance under any noise level at test time and lower than performance if trained without noise. Our work starts exactly here, aiming to study how noise affects the learning dynamics.

\paragraph{RL that matters.}
To mitigate the known and well documented brittleness of RL \citep{henderson-aaai18a,engstrom-iclr20a,andrychowicz-iclr21a}, various research directions are actively being explored. Under the umbrella-term AutoRL, \citet{parkerholder-jair22a} identify multiple dimensions that are (automatically) adapted to improve robustness of learning pipelines. For example, various methods and techniques are proposed to automatically adapt the hyperparameters of RL pipelines, to increase the final performance \citep[see, e.g., ][]{jaderberg-arxiv17a,franke-iclr21a,parkerholder-neurips20a}, as well as improve robustness \citep{falkner-icml18a,eimer-icml23a}.
Further, environment design approaches propose to augment different elements of the environment, including reward function augmentation \citep{marthi-icml07a,faust-automl19a,hu-neurips20a}, action spaces \citep{mankowitz-aaai18a,farahani-asc19a,bagaria-ijcai21a} as well as observation space augmentation \citep{raileanu-neurips21a,yarats-iclr21a}. The latter is of particular interest to our study as such work considers automatic selection of transformations (such as crop, rotate and flip) to image observations, such that policies are invariant to these transformations while maximizing rewards.
In contrast to these approaches, which primarily prioritize maximizing final rewards, our study shifts the focus toward deepening the understanding of reinforcement learning in noisy environments and exploring how plasticity-preserving methods can enhance robustness under such conditions. Thus, we focus on settings in which we draw random samples, rather than randomly perturbing samples.

\section{Background}
\subsection{MDPs, RL and the Effect of Different Sources of Noise} \label{sec:background_noisy_mdps}

The goal of \gls*{rl} is to find the optimal policy in a \gls*{mdp}.
An \gls*{mdp} is defined by a tuple $\gM = \left( \gS, \gA, P, r, \gamma  \right)$ where $\gS$ is the state space, $\gA$ is the action space, $P: \gS \times \gA \times \gS \rightarrow [0, 1]$ 
defines the transition probability distribution, $r: \gS \times \gA \rightarrow \sR$
is the reward function and $\gamma \in [0,1]$ is the discount factor.
An optimal policy
$\pi: \gS \times \gA \rightarrow [0, 1]$ 
then needs to maximize the expected discounted return 
$\E_{\tau(\pi)} \left[ \sum_{t=0}^{\infty} \gamma^t r(s_t,a_t) \right]$.

Perturbing the observations of the agent at time $t$ as $\tilde{\ro}_t \sim \nu(\cdot | s_t)$ breaks the Markov property and gives rise to a \gls*{pomdp} $\gM' = \left( \gS, \gA, P, r, \gamma, \nu, \gO \right)$ where $\gO$ is the observation space. Hence, solving the \gls*{pomdp} generally involves a state estimation problem to maintain a belief over the true state \citep{kaelbling-ai98a}.

In contrast, perturbations on the actions as $\tilde{\ra}_t \sim \rho(\cdot | a_t)$ yield a transformed \gls*{mdp} with a transition probability function which ``absorbs'' the action noise:
\begin{equation} \label{eq:transition_anoise}
  P'(s_{t+1}|s_t,a_t) = \int_\gA P(s_{t+1}|s_t,\tilde{a}) \rho(\tilde{a}|a_t)d\tilde{a}
\end{equation}

\subsection{Quantifying Plasticity Loss}
The loss of plasticity of a neural network describes its decreasing ability to fit a changing data distribution or to ``respond to different \textit{possible} learning signals'' \citep{lyle-icml23a}.
\citet{lyle-icml23a} propose to quantify it as the ability of a (trained) neural network parameterized by $\theta_t$ to fit a high frequency transformation of the output of a randomly initialized model from the same class: $y(x) = b + \sin\left(M \cdot f_{\theta_0}(x)\right)$, where $M=10^5$ and $b$ is a bias to compensate for the offset of the mean prediction of $f_{\theta_t}$. 
This definition focuses on the ability to reduce the \textit{training} loss, whereas other work has identified similar phenomena in terms of \textit{generalization} capabilities of neural networks \citep{ash-neurips20a,igl-iclr21a}.

A broad range of work explains plasticity loss by a reduced representational and signal propagation capacity when compared to the time of initialization. Prominent measures to detect such capacity loss are dormant neurons \citep{sokar-icml23a}, a low effective rank of feature representations \citep{kumar-iclr21a} or linearization of activation functions \citep{lyle-collas24a}. Further, optimization pathologies such as loss landscape sharpness and a degenerate neural-tangent kernel \citep{lyle-collas24a} have been connected to the cause of plasticity loss.

Previous work suggests that preserving the plasticity of the critic is more important in \gls*{sac} \citep{ma-iclr24a,nauman-icml24a}.
This likely is due to the fact that the phenomena treated under the term plasticity loss have also been studied more specifically in the context of \gls*{td} learning.
\citet{kumar-iclr21a} studied and explained the decrease of the effective rank in deep \gls*{rl} as a direct result of \gls*{td} learning.
And \citet{hussing-rlj24a} attributed the primacy-bias not to overfitting on early data but to early value divergence due to limited action-space coverage of samples at the beginning of training.

We will refer to $Q$-value divergence as \gls*{td} instability as one specific mechanism that may cause severe plasticity loss, in accordance with the findings by \citet{lyle-collas24a} that regression on high offset targets results in particularly severe loss of plasticity.

\section{Maintaining Comparable Plasticity Probing Throughout Training}

To assess the ability of a neural network implementing a parametric function $f_\theta: \sR^n \to \sR^m$ to learn from non-stationary data which was gathered in a stochastic environment, we aim to directly quantify the plasticity of its parameters $\theta$.
To this end we adopt the definition of plasticity as proposed by \citet{lyle-icml23a}:

\begin{definition}[Optimization procedure] \label{def:optimization_procedure}
  Given dataset $\mD$ and a loss function $L(\cdot \mid \mD)$, we define an optimization (or update) procedure $\gU: \Theta \times \sN_+ \rightarrow \Theta$ as a function that returns updated parameters $\theta' = \gU(\theta, T)$, by performing $T \in \sN_+$ steps minimizing $L(\cdot \mid \mD)$ starting from $\theta \in \Theta$.
\end{definition}

\begin{wrapfigure}{r}{0.4\textwidth}\vskip -1.5em
  \centering
  \includegraphics[width=0.38\textwidth]{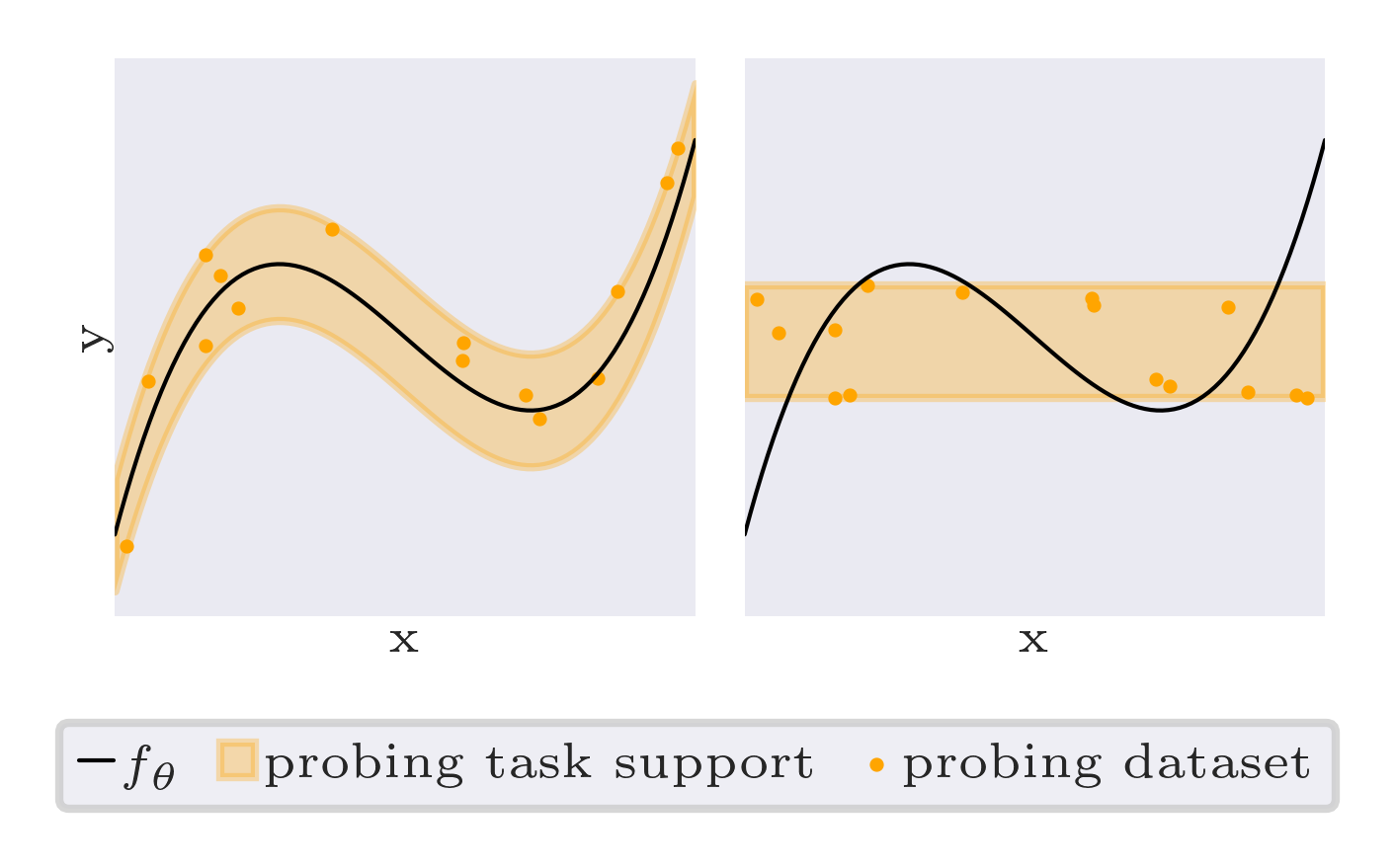}
  \caption{Illustration of the probing task. As implemented in this study (left); as originally proposed by \citet{lyle-icml23a} (right).}\label{fig:probing_task}
\end{wrapfigure}

Together with a neural network architecture, $\theta$ implements the function $f_\theta$. The optimization procedure $\gU$ may have further updatable parameters, such as dynamic learning rates or gradient moment statistics. For the sake of a clear presentation, we subsume these parameters into $\theta$ and note that they have no immediate effect on the computation of $L(\theta \mid \mD)$\footnote{
  However, these optimizer statistics can have a significant impact on adaptation to distribution shifts and hence plasticity loss \citep{lyle-icml23a}.
  }.
  Likewise the optimization procedure is implicitly specified by the choice of the loss function $L$ and the dataset $\mD$ and we omit the explicit dependence for the sake of brevity.

\begin{definition}[Neural network plasticity]
Given an optimization procedure $\gU$ as in \cref{def:optimization_procedure}, a neural network parameterized by $\theta$, a probing dataset $\mD$ and a number of probing steps $T$ we define plasticity as:

  \begin{equation} \label{eq:plasticity}
    \gP(\theta | \mD) = \frac{L(\theta | \mD) - L\left(\gU(\theta, T)|\mD\right)}{L(\theta | \mD)}
  \end{equation}
  
\end{definition}
That is, plasticity is the fraction of the initial loss $L\left(\theta | \mD \right)$ that can be removed by the update procedure $\gU$ on the task $L\left(\cdot | \mD \right)$ within $T$ probing steps.

Following \citet{lyle-icml23a}, we define the probing task as an MSE-regression task, however with an important distinction:
We generate the probing task by adding randomly-sampled and sine-transformed offsets to the current network's predictions.
I.e., given a sample $\vx$ we set the target as $\vy = f_\theta(\vx) + \sin\left( \mathbf{\epsilon} \right)$, where $\mathbf{\epsilon} \sim \Pi_{i=1}^m \text{Unif}\left([0, 2 \pi]\right)$ is sampled from a uniform distribution and has the same dimensionality as the output of $f_\theta$.

The central difference to \citet{lyle-icml23a} is that we offset $\vy$ of the probing samples specific to the respective input $\vx$ instead of using a global offset $\vb$, which they set to the mean prediction of $f_\theta$ over $\mD$. 
By considering the ``shape'' of $f_\theta$ we keep the initial probing loss $L\left(\theta | \mD\right)$ approximately constant since all probing samples lie within an $\epsilon$-ball around $f_\theta$. We illustrate the difference between the original and our approach in \Cref{fig:probing_task}. We can think of this as probing the ability of the model to fit perturbations in arbitrary directions.

\section{Experiments}

\subsection{Setup and Evaluation Protocol}
\paragraph{Stochastic environments.}
We study additive observation and action noise for two reasons:
(1) both can be implemented by wrapping existing \gls*{dmc} environments (humanoid-walk, quadruped-run and walker-run) without modifying the physics simulation, making the approach straightforward to extend to other benchmarks \citep{rajan-jmlr23a};
(2) they affect learning in fundamentally different ways: observation noise breaks the Markov property and gives rise to a \gls*{pomdp}, whereas action noise induces a transformed \gls*{mdp} that adds stochasticity to otherwise deterministic dynamics (Section~\ref{sec:background_noisy_mdps}).
Following \citet{rajan-jmlr23a}, the wrapper transforms the agent's action before passing it to the environment as $\tilde{\va}_t = \va_t + \epsilon_{\va, t},\ \epsilon_{\va, t} \sim \gN\left(0,\sigma_\va\right)$ and the next-state observation to $\tilde{\vo}_{t+1} = \vo_{t+1} + \epsilon_{\vo, t},\ \epsilon_{\vo, t} \sim \gN\left(0,\sigma_\vo\right)$ before returning it to the agent.\footnote{To facilitate better reproducibility of our research, we open-source our code and research artifacts.
The code to reproduce our experiments, as well as links to the generated data, can be accessed via \url{https://github.com/PhilippBordne/noisy-plasticity}.}

\paragraph{Learning algorithm.}
We train \gls*{sac} agents \citep{haarnoja-icml18a,haarnoja-arxiv19a} for one million environment steps with default hyperparameters (see Appendix~\ref{app:hp_select_sac}).
We vary the replay ratio (gradient steps per environment step) over $\{1, 4, 8\}$ to investigate whether higher update frequencies exacerbate overfitting under noise \citep{nikishin-icml22a,igl-iclr21a}.
When adding \gls*{ln} or resets we follow the implementation details of the respective papers \citep{nikishin-icml22a,lyle-icml23a,lyle-collas24a}: \gls*{ln} is inserted before each ReLU activation and resets are applied to all training networks (actor, critics) and Adam optimizer statistics every $200k$ steps.

\paragraph{Evaluation protocol.}
We report performance as the mean over 5 evaluation episodes in a separately seeded evaluation environment.
Increasing noise levels reduce the performance attainable by any method (Figure~\ref{fig:simple_baselines_across_noise}), compressing the range of achievable scores toward the random-agent baseline.
As a result, we deem an equal absolute performance gap between two methods to carry more weight under high noise than in a noise-free setting.
To account for this, we normalize scores per task and noise level using the Normalized Score of \citet{agarwal-neurips21a}:

\begin{equation}\label{eq:normalized_score}
    \text{Normalized Score} = \frac{\text{Score} - \text{Minimal Score}}{\text{Target Score} - \text{Minimal Score}}
\end{equation}

In absence of a clearly defined reference performance, such as a game solved in ATARI, we set the target score as the mean over the best configuration per task and training regime, and the minimal score is given by the performance of a random agent in the environment.
This normalization ensures that method comparisons are expressed relative to the difficulty of each noise condition, making differences across noise levels and tasks commensurable.
For reporting per method we select the best replay ratio based on mean training performance over the last 5000 environment steps, decoupling hyperparameter selection from evaluation. We report the raw learning curves in Appendix~\ref{app:learning_curves_complete}.

\subsection{The Effect of Noise During Training on SAC}
\begin{figure}[h]
  \centering
  \includegraphics[width=\textwidth]{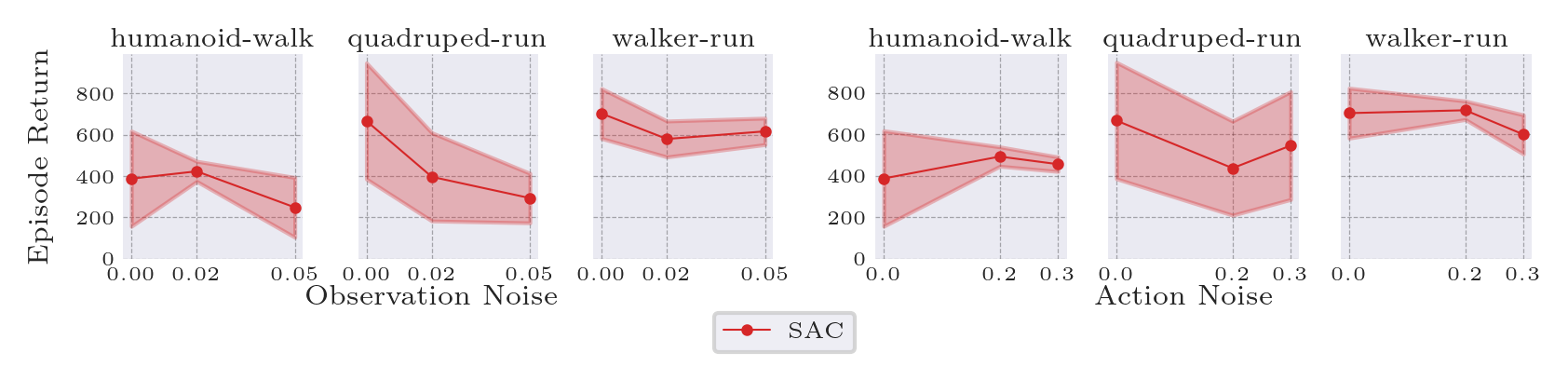}
  \caption{Final performance of \gls*{sac} under increasing observation noise (left) and action noise (right) after $10^6$ training steps.} \label{fig:simple_baselines_across_noise}
\end{figure}

\citet{glossop-neurips22a} show that even low levels of observation noise \textit{during training} are far more detrimental than the same noise applied only at evaluation time, suggesting that noise actively disrupts the learning process rather than merely making the task harder to execute.
We confirm this finding on proprioceptive control tasks from \gls*{dmc}. Figure~\ref{fig:simple_baselines_across_noise} shows that \gls*{sac} is already strongly impaired by mild observation noise on the challenging tasks \texttt{quadruped-run} and \texttt{humanoid-walk}, whereas action noise causes a noticeable drop only at higher magnitudes.

Note that noise magnitudes across types are not directly comparable. The action noise is first absorbed by the environment's dynamics before affecting the next state, so its effective impact on the state distribution can differ substantially from the nominal $\sigma_\va$ (see Appendix~\ref{app:noise_on_dynamics}).
This makes it difficult to attribute the stronger impact of observation noise unambiguously to the Markov property violation versus the perturbation of observation dimensions critical for action selection.

\subsection{Can Plasticity-Preserving Methods Improve Training Robustness Under Noise?} \label{sec:robustness_plasticity_methods}

Following the motivation laid out in \cref{sec:introduction}, we ask whether methods proposed to preserve plasticity and counteract overfitting to early transitions \citep{nikishin-icml22a,igl-iclr21a} also improve robustness when learning under noise.

Two interventions stand out as particularly effective on the \gls*{dmc} \citep{nauman-icml24a}, each targeting a distinct mechanism.
\gls*{ln} \citep{ba-arxiv16a} has been motivated from two complementary angles. On the one hand, \citet{lyle-collas24a} show that distribution shifts can cause large pre-activation changes that push neurons into saturation and thereby reduce plasticity, and that \gls*{ln} mitigates this. On the other hand, constraining feature activation norms stabilizes \gls*{td} learning \citep{bhatt-iclr24a,gallici-iclr25a,hussing-rlj24a}.
Periodic parameter resets \citep{nikishin-icml22a} directly counteract overfitting to early samples by periodically reinitializing network weights, erasing the temporal ordering bias accumulated in the replay buffer \citep{igl-iclr21a}.

\begin{figure}[ht]
    \centering
    \includegraphics[width=\textwidth]{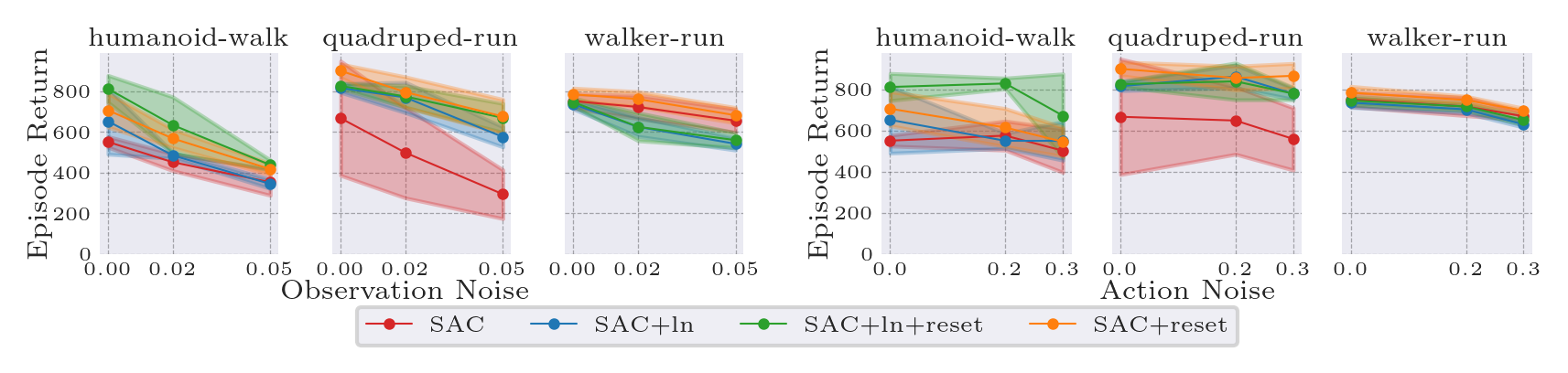}
    \caption{Episode returns from final evaluation after $10^6$ steps. Each point aggregates to mean and std over 5 seeds per method and its best replay ratio.}\label{fig:abs_performance_per_env}
\end{figure}

\begin{figure}
    \centering
    \includegraphics[width=\textwidth]{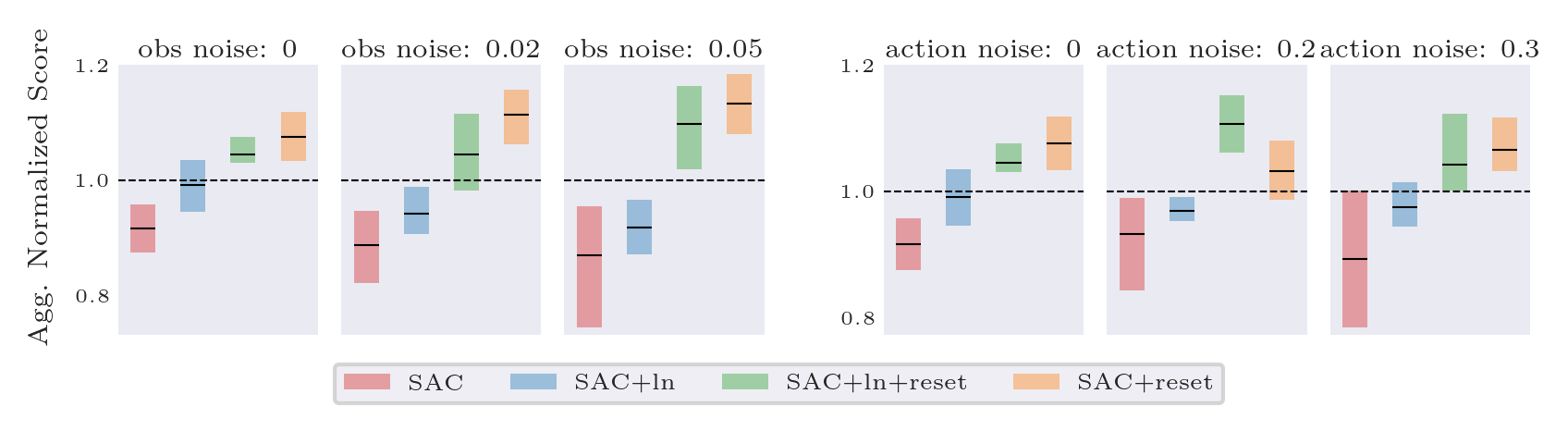}
    \caption{Aggregated normalized scores (stratified bootstrapping, \gls*{iqm} with 95\% CI \citep{agarwal-neurips21a}) across environments. Per method, environment and noise level the results for the best replay ratio were selected for aggregation. \textit{Left} (observation noise): the relative effectiveness of periodic resets grows with noise magnitude; resets alone consistently outperform resets combined with \gls*{ln}, while baseline \gls*{sac} and \gls*{ln} alone deteriorate at higher noise levels. \textit{Right} (action noise): no consistent ranking emerges and the relative performance of \gls*{sac} becomes increasingly uncertain.}
    \label{fig:agg_best_per_noise}
\end{figure}

Adding \gls*{ln}, periodic resets, or both improves final performance on the two harder tasks but not on \texttt{walker-run} (Figure~\ref{fig:abs_performance_per_env}), with the largest gains on \texttt{quadruped-run} where baseline \gls*{sac} degrades rapidly under observation noise.
On \texttt{humanoid-walk}, however, all four configurations converge at $\sigma_\vo = 0.05$.
Aggregated normalized scores (Figure~\ref{fig:agg_best_per_noise}, left) confirm that the relative advantage of periodic resets grows with increasing observation noise.
Resetting alone outperforms resets combined with \gls*{ln} at all three observation noise levels, while baseline \gls*{sac} and \gls*{ln} alone fare progressively worse.
Under action noise (Figure~\ref{fig:agg_best_per_noise}, right), reset-based methods retain their advantage over baseline \gls*{sac}, but it no longer grows with the noise level and the ranking between the two reset variants flips.

\subsection{Can Plasticity Preservation Explain Robustness Under Noise?}

The results in Section~\ref{sec:robustness_plasticity_methods} show that plasticity-preserving methods improve robustness under noise, but cannot fully prevent performance degradation.
This is unsurprising in part as observation noise turns the \gls*{mdp} into a \gls*{pomdp} and thus some performance loss is inevitable regardless of plasticity.
Beyond this, we use our results to better understand how noise and plasticity loss interact during learning. Specifically, we ask whether this relationship is unidirectional, with noise impairing plasticity which then harms performance, or whether learning failure and plasticity loss mutually reinforce each other.

\begin{figure}[h]
  \centering
  \includegraphics[width=\textwidth]{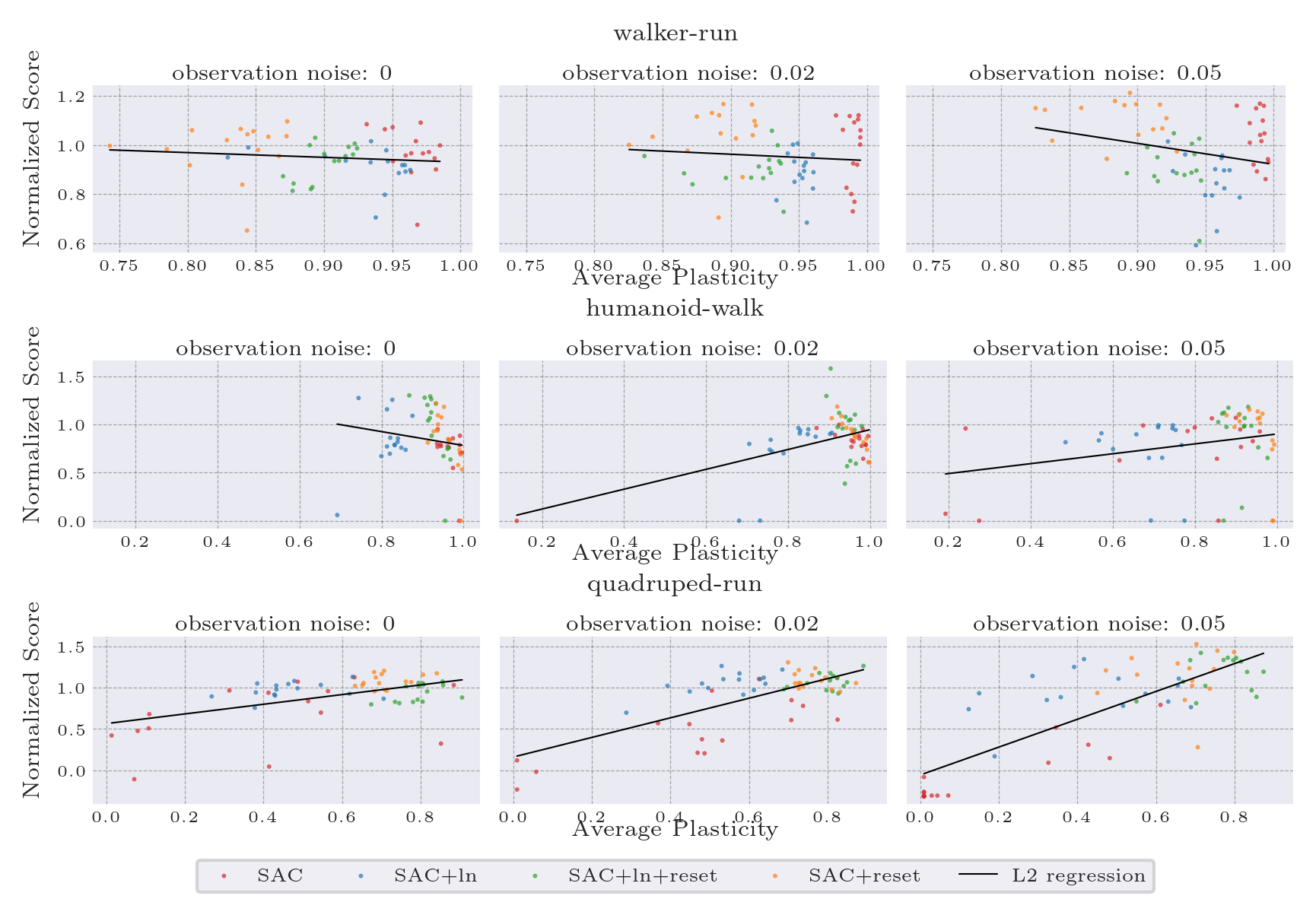}
  \caption{Mean critic plasticity over training vs.\ final normalized score, per environment and observation noise level ($\sigma_\text{o}$). Per plot, an individual point is a combination of (method, seed, replay ratio). An $L2$-regression line is fit across all runs per plot.} \label{fig:return_vs_plasticity_onoise}
\end{figure}

To this end, Figure~\ref{fig:return_vs_plasticity_onoise} plots mean critic plasticity across training against final normalized score, for each combination of method, environment, and observation noise level.
We use mean plasticity as a proxy for the agent's sustained ability to adapt its representations in response to the learning signal.\footnote{Plasticity as defined in \eqref{eq:plasticity} can occasionally be negative due to destructive gradient interference across probing batches; we clip values to $[0, 1]$ before averaging.}

\paragraph{Hard tasks: noise and plasticity loss reinforce each other.}
On \texttt{quadruped-run}, most runs that fail to improve over the random baseline show a catastrophic loss of plasticity, whereas runs that retain higher levels of plasticity tend to reach a better final return (Figure~\ref{fig:return_vs_plasticity_onoise}).
Crucially, higher noise also shifts the distribution of learners toward low-plasticity outcomes, indicating the relationship is not unidirectional.
Noise does not merely make the task harder for an otherwise intact agent; it also accelerates plasticity loss, which then further undermines learning.
On \texttt{humanoid-walk} the same trend appears but is less pronounced. Agents with only mild plasticity loss frequently fail under high noise, and some failure is already observed in the noise-free setting, which shows that plasticity loss is only one of several factors making this particular task difficult.

\paragraph{Easy tasks: noise can prevent overfitting.}
Compared to the previous observations, \texttt{walker-run} shows the opposite pattern. Adding observation noise leads to \textit{more} plastic learners on average, and final performance is weakly negatively correlated with plasticity.
We hypothesize that noise acts as a mild regularizer in this setting. By making the task slightly harder, it prevents early convergence to simple behaviors that exhibit reduced plasticity.

\paragraph{Action noise.}
We observe qualitatively similar trends under the highest investigated level of action noise ($\sigma_\va = 0.3$), but not at lower magnitudes, consistent with the smaller absolute performance drops seen at those levels (Figure~\ref{fig:abs_performance_per_env}; see also Appendix~\ref{app:plasticity_vs_robustness_anoise}).
These findings indicate that an environment's stochasticity itself is a factor reducing learning robustness, and that this cannot be traced back to partial observability alone.

\subsection{When Does Plasticity Loss Occur and Are Early Dynamics Critical?}
\begin{figure}
  \centering
  \includegraphics[width=\textwidth]{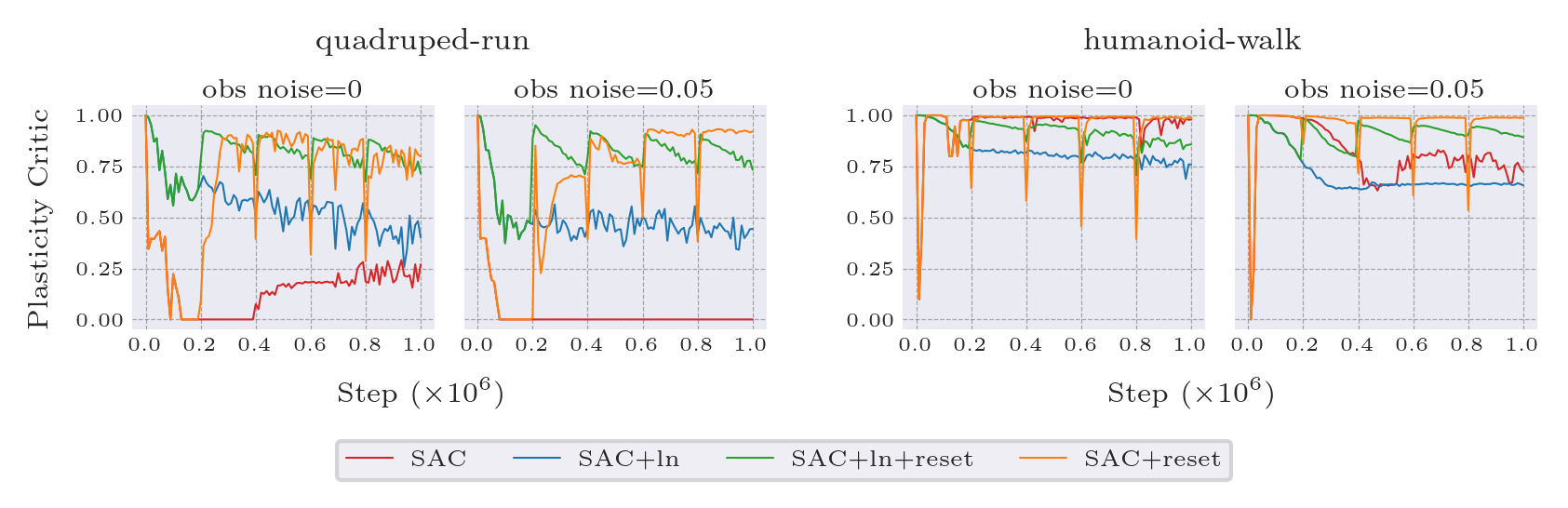}
  \caption{Evolution of critic plasticity throughout training on the \texttt{quadruped-run} and \texttt{humanoid-walk} tasks. Each curve shows the mean over 5 seeds. Plasticity values have been clipped to $[0,1]$ before averaging.}\label{fig:plasticity_snapshots}
\end{figure}

In Section~\ref{sec:robustness_plasticity_methods} we motivated periodic resets as a remedy for overfitting to the noise-corrupted early transitions that dominate the replay buffer.
This raises a more specific question: ``During which training phase does plasticity loss actually occur, and is the early phase particularly critical?''
If so, this would motivate more targeted interventions that apply resets aggressively early but less frequently later, trading plasticity preservation for learning speed.

To investigate, we focus on the most challenging settings in our study, \texttt{quadruped-run} and \texttt{humanoid-walk} under high observation noise ($\sigma_\vo = 0.05$), and plot mean critic plasticity throughout training (Figure~\ref{fig:plasticity_snapshots}).
We use a replay ratio of 4 to amplify any effects on plasticity and \gls*{td} stability.

Consistent with prior work \citep{hussing-rlj24a}, plasticity loss is most pronounced in the very early stages of training and is further exacerbated by observation noise (Figure~\ref{fig:plasticity_snapshots}).
For methods with periodic resets, the plasticity drop following each subsequent reset becomes progressively less severe.
Notably, combining resets with \gls*{ln} does not fully eliminate this pattern: a persistent plasticity drop follows each reset, whereas resets alone eventually yield a regime with negligible plasticity loss between cycles.

\begin{figure}
  \centering
  \includegraphics[width=\textwidth]{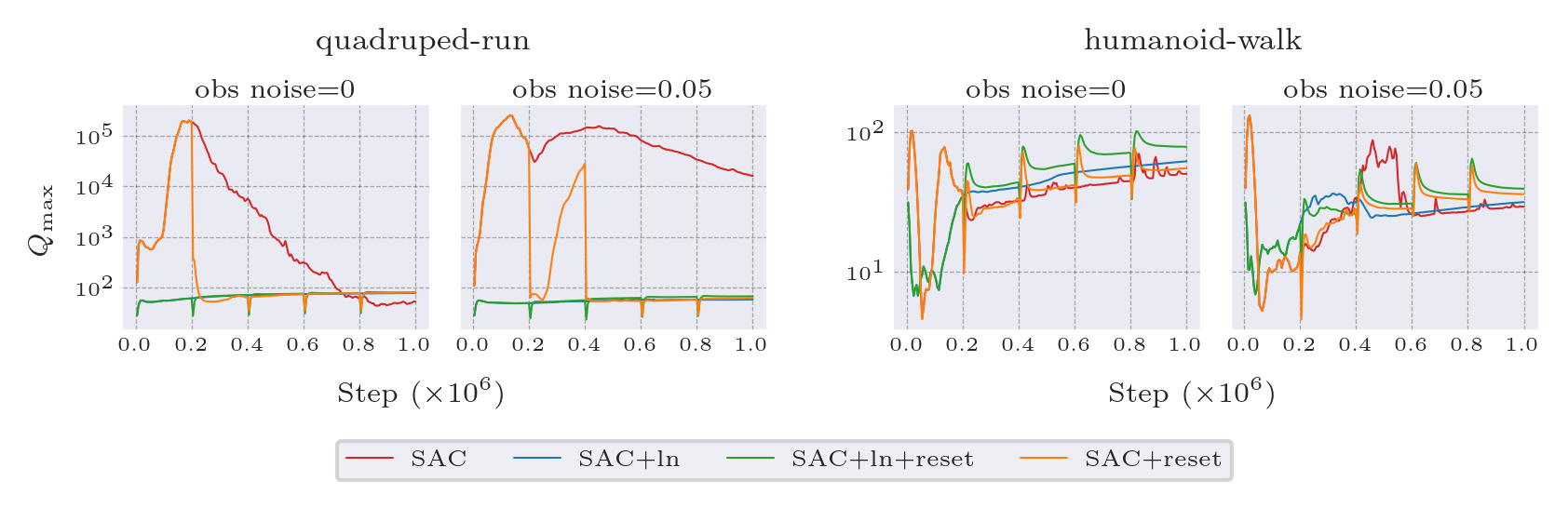}
  \caption{Evolution of the maximum $Q$-value within training batches throughout training on the \texttt{quadruped-run} and \texttt{humanoid-walk} tasks. Each curve shows the mean over 5 seeds.}\label{fig:qmax_snapshots}
\end{figure}

To further distinguish plasticity loss from \gls*{td} instability, which \citet{hussing-rlj24a} identified as the primary failure mode, we track the maximum $Q$-value within training batches (Figure~\ref{fig:qmax_snapshots}).
For baseline \gls*{sac} on \texttt{quadruped-run}, $Q$-values diverge dramatically under high observation noise, indicating that TD updates become unstable.
On \texttt{humanoid-walk} no comparable overestimation occurs, consistent with the smaller performance differences between methods seen on that task.
These results suggest that in stochastic environments, \gls*{td} instability is a particularly important failure mode to prevent. While \gls*{td} instability is catastrophic in any setting, noise makes it more likely to occur whereas plasticity loss and overfitting to early samples alone tend to cause only gradual drops in performance.

\section{Limitations}
Due to the computational overhead of evaluation over different noise types and levels we had to limit our experiments to a subset of benchmarking environments.
Nevertheless, in terms of the evaluation setting we see a bigger limitation in the characteristics of the studied noise types, in particular applying the same level of noise indiscriminately of the observation dimension.
We also believe that studying noisy transition dynamics is of great interest to better understand the interplay between plasticity loss and potential overfitting to early, noisy samples without directly transforming the setting from an \gls*{mdp} into a \gls*{pomdp}.
We expect its implementation to be more difficult in relative terms as it will be necessary to implement this on the simulation level and can not be achieved by a simple wrapper.
In line with prior works, we opted to apply unrealistically high levels of action noise. However, \citet{glossop-neurips22a} suggest that true transition noise would result in a noticeable impact earlier.
Lastly, while plasticity loss and \gls*{td} stability have been established as important factors for the successful application of deep \gls*{rl}, they are only two of many and studying their interplay with other aspects, such as exploration, is required to gain a holistic understanding of the effectiveness of the evaluated methods.

\section{Conclusion}
Aggregated across the three \gls*{dmc} tasks studied, we find that plasticity-preserving methods improve learning robustness in stochastic environments: reset-based agents achieve a higher aggregate normalized score than baseline \gls*{sac} at every noise type and level tested.
Under observation noise this advantage grows with the noise level, meeting the criterion set out in \cref{sec:introduction}, whereas under action noise it is persistent but independent of the noise level.
Moreover, we observe an increased loss of plasticity in these settings, and in the most drastic case (baseline \gls*{sac} on \texttt{quadruped-run} under $\sigma_\vo = 0.05$) observation noise led to severe value overestimation from which training did not recover.
Since stochasticity makes environments inherently more difficult, it is unsurprising that the methods investigated can mitigate but not fully prevent the decrease in performance.
Overall, our findings suggest that for \gls*{sac} in stochastic environments, maintaining \gls*{td} stability is more critical than preventing overfitting to perturbed, early samples.
We found \gls*{ln} to be primarily effective in preventing value divergence while permitting some loss of plasticity.
Under observation noise, resetting was the most effective of the four configurations tested, providing both long-term \gls*{td} stability and plasticity preservation.
Somewhat surprisingly, combining resetting with \gls*{ln} underperformed resetting alone at every observation noise level tested and led to an ongoing, albeit slow, loss of plasticity throughout training.

We should expect data collected in the real world to be stochastic by nature and hence we identify a need for benchmark environments that allow the assessment of robust learning under such conditions for the holistic assessment of an algorithm's capabilities.
While we resorted to a simple extension of existing benchmark environments we envision benchmarks with more realistic non-deterministic characteristics that we should expect in the real world such as stochastic environment dynamics, random observation delays or observation noise that is well-calibrated with regard to the overall signal magnitude on the respective channels.
Once access to such simulators exists, we should aim to further understand the interplay between environment stochasticity, plasticity loss and further central components of \gls*{rl} algorithms.
Lastly, evaluation in such environments should inform the further development and refinement of algorithms in general and plasticity preserving methods in particular.

\begin{ack}
The authors acknowledge support by the state of Baden - Württemberg through bwHPC and the German Research Foundation (DFG) through grant INST 35/1597-1 FUGG.
The authors further acknowledge funding from The Carl Zeiss Foundation through the research network ``Responsive and Scalable Learning for Robots Assisting Humans'' (ReScaLe) of the University of Freiburg.
This research was funded by the Deutsche Forschungsgemeinschaft (DFG, German Research Foundation) under grant number 539134284, through EFRE (FEIH\_2698644) and the state of Baden-Württemberg.
\begin{center}
  \includegraphics[width=0.3\textwidth]{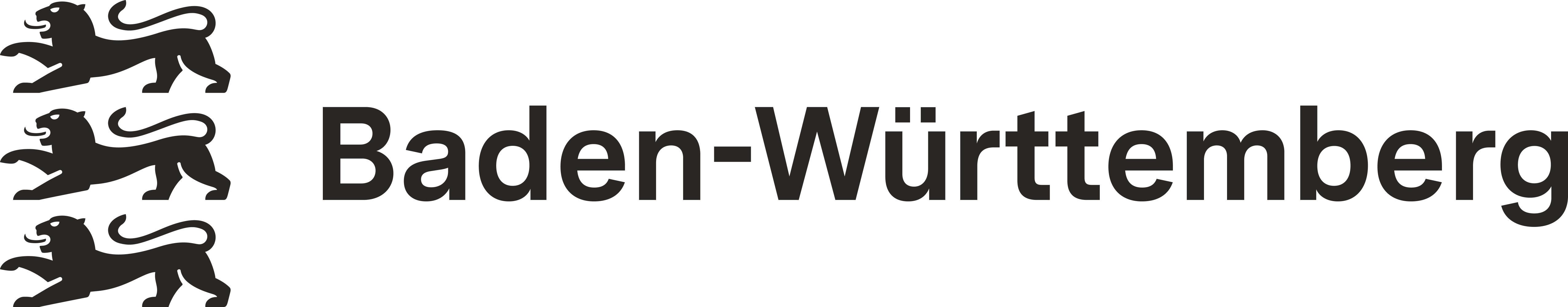} ~~~
  \includegraphics[width=0.3\textwidth]{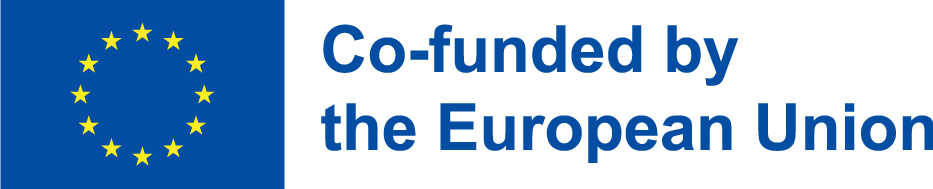}
\end{center}
\end{ack}

\bibliographystyle{plainnat}
\bibliography{plastic_lib}

\clearpage
\appendix
\section{Hyperparameter Settings Soft Actor Critic} \label{app:hp_select_sac}
For our study we use the SAC implementation of CleanRL \citep{huang-jmlr22a}. We keep the default hyperparameters and only vary the replay ratio. We report the hyperparameters in \Cref{tab:fixed_hp_sac}. Further details about hyperparameters can be seen in our code repository.
\begin{table}[h!]%
  \centering
  \caption{Hyperparameters for SAC.}\label{tab:fixed_hp_sac}
  \begin{tabular}{|l|l|}
    \hline
    optimizer & Adam \\
    \hline
    batch size & 256 \\
    \hline
    $\beta_1$ (Adam) & 0.9 \\
    \hline
    $\beta_2$ (Adam) & 0.999 \\
    \hline
    learning rate & \texttt{3e-4} \\
    \hline
    buffer size & \texttt{1e6} \\
    \hline
    $\gamma$ (discount factor) & 0.99 \\
    \hline
    $\tau$ (target smoothing coefficient) & 0.005 \\
    \hline
    $\alpha$ (entropy coefficient, fixed) & 0.2 \\
    \hline
    train frequency & 1 \\
    \hline
    target network frequency & 1 \\
    \hline
    replay ratio & [1, 4, 8] \\
    \hline
    learning starts & 1000 \\
    \hline
  \end{tabular}
\end{table}

\clearpage
\section{Learning Curves and Normalized Scores per Environment} \label{app:learning_curves_complete}
\begin{figure}[hb!]
  \centering
  \includegraphics[width=\textwidth]{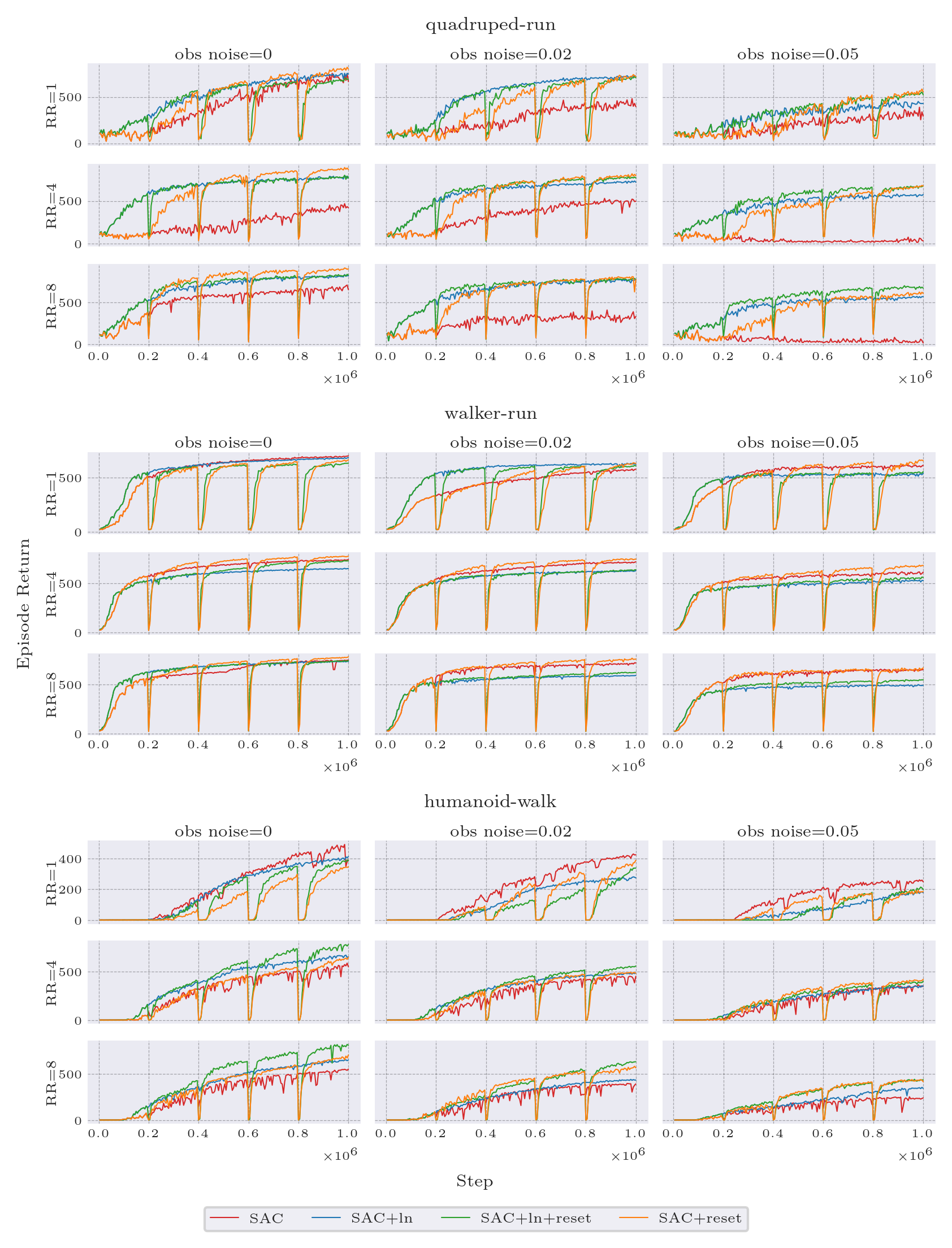}\caption{Learning curves for the evaluation episode return across replay ratios and observation noise levels.}\label{fig:app_rr_lc_on}
\end{figure}

\begin{figure}[hb!]
  \centering
  \includegraphics[width=\textwidth]{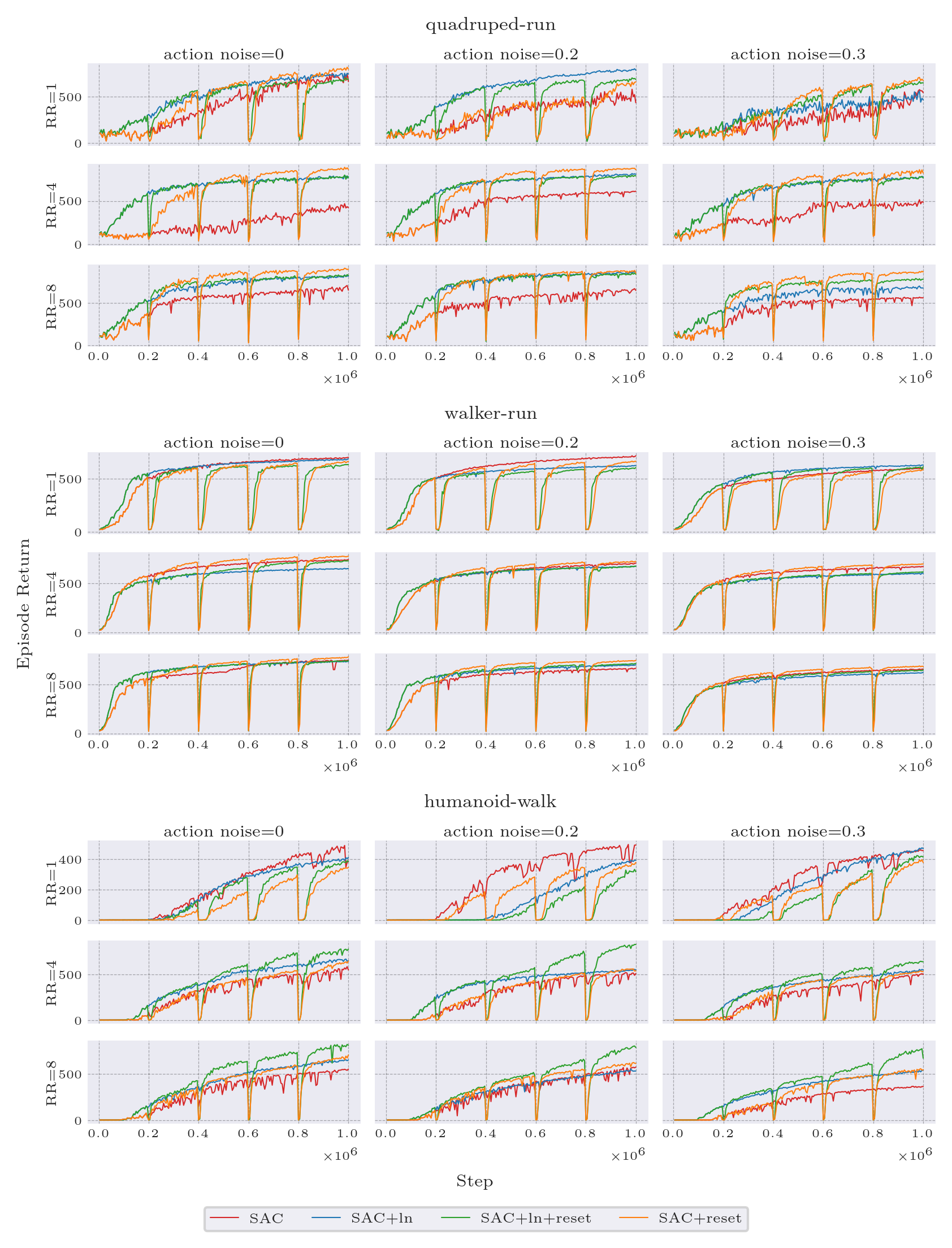}\caption{Learning curves for the evaluation episode return across replay ratios and action noise levels.}\label{fig:app_rr_lc_an}
\end{figure}

\begin{figure}[hb!]
  \centering
  \includegraphics[width=\textwidth]{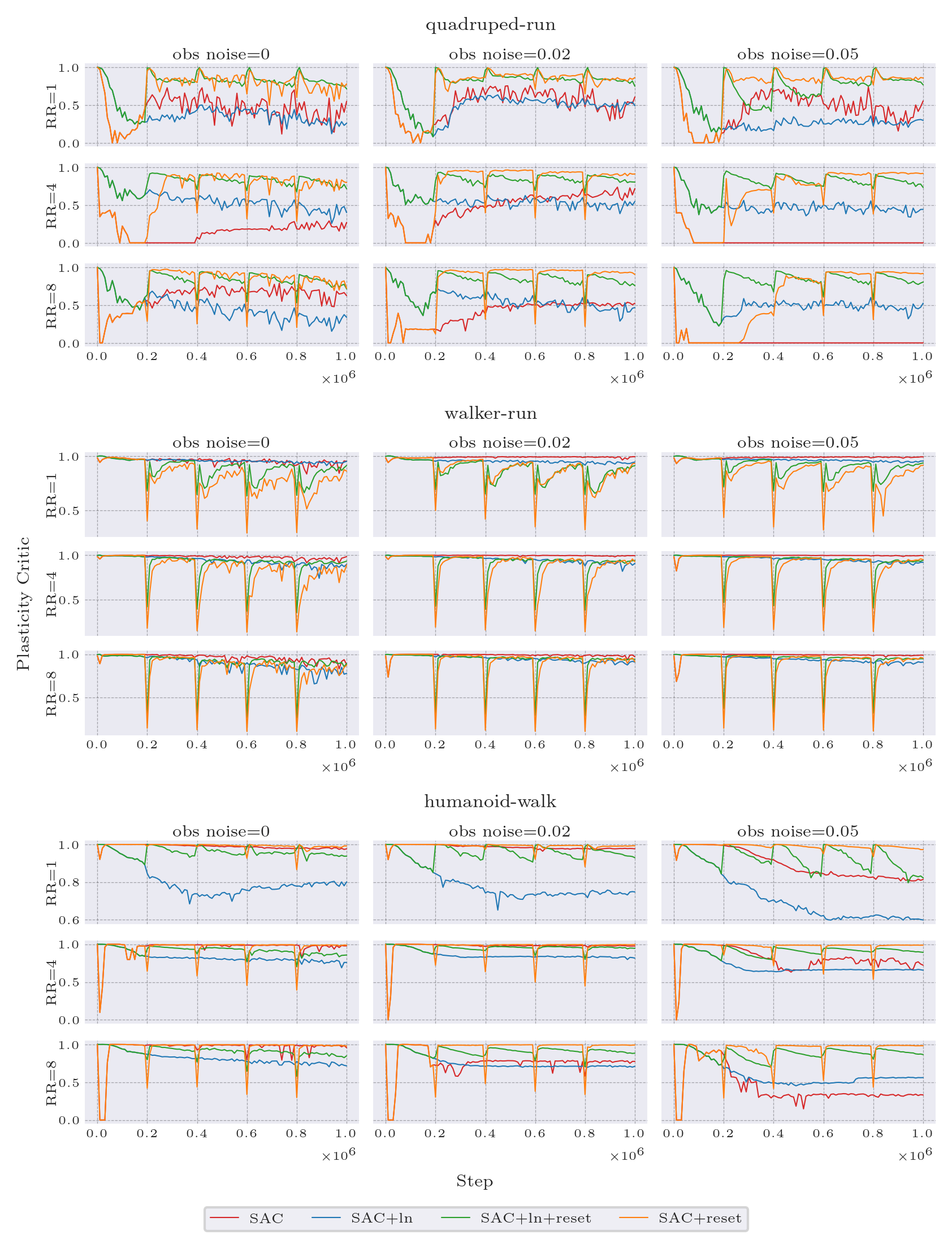}\caption{Learning curves for the plasticity evolution across replay ratios and observation noise levels.}\label{fig:app_pe_lc_on}
\end{figure}

\begin{figure}[hb!]
  \centering
  \includegraphics[width=\textwidth]{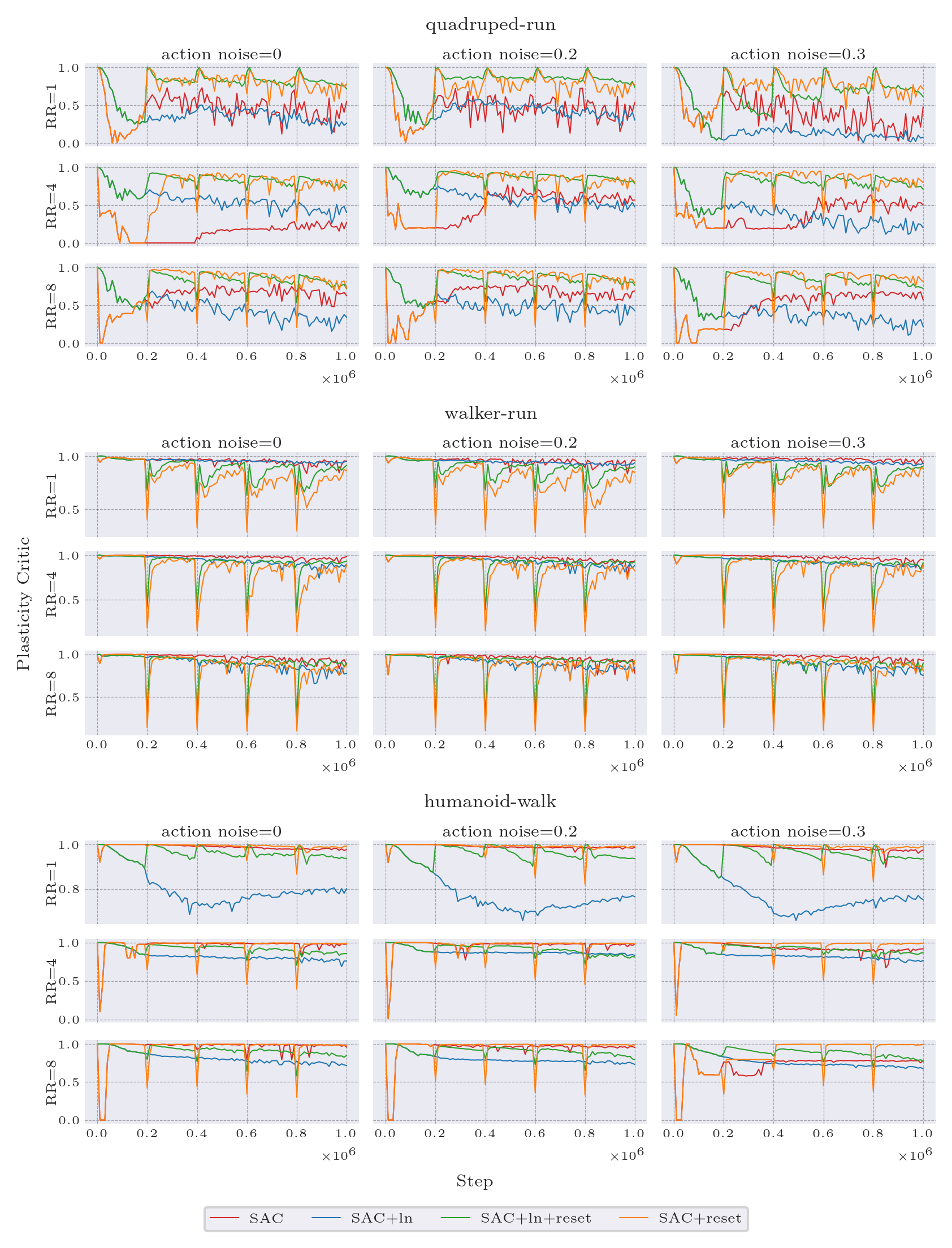}\caption{Learning curves for the plasticity evolution across replay ratios and action noise levels.}\label{fig:app_pe_lc_an}
\end{figure}

\begin{figure}[hb!]
  \centering
  \includegraphics[width=\textwidth]{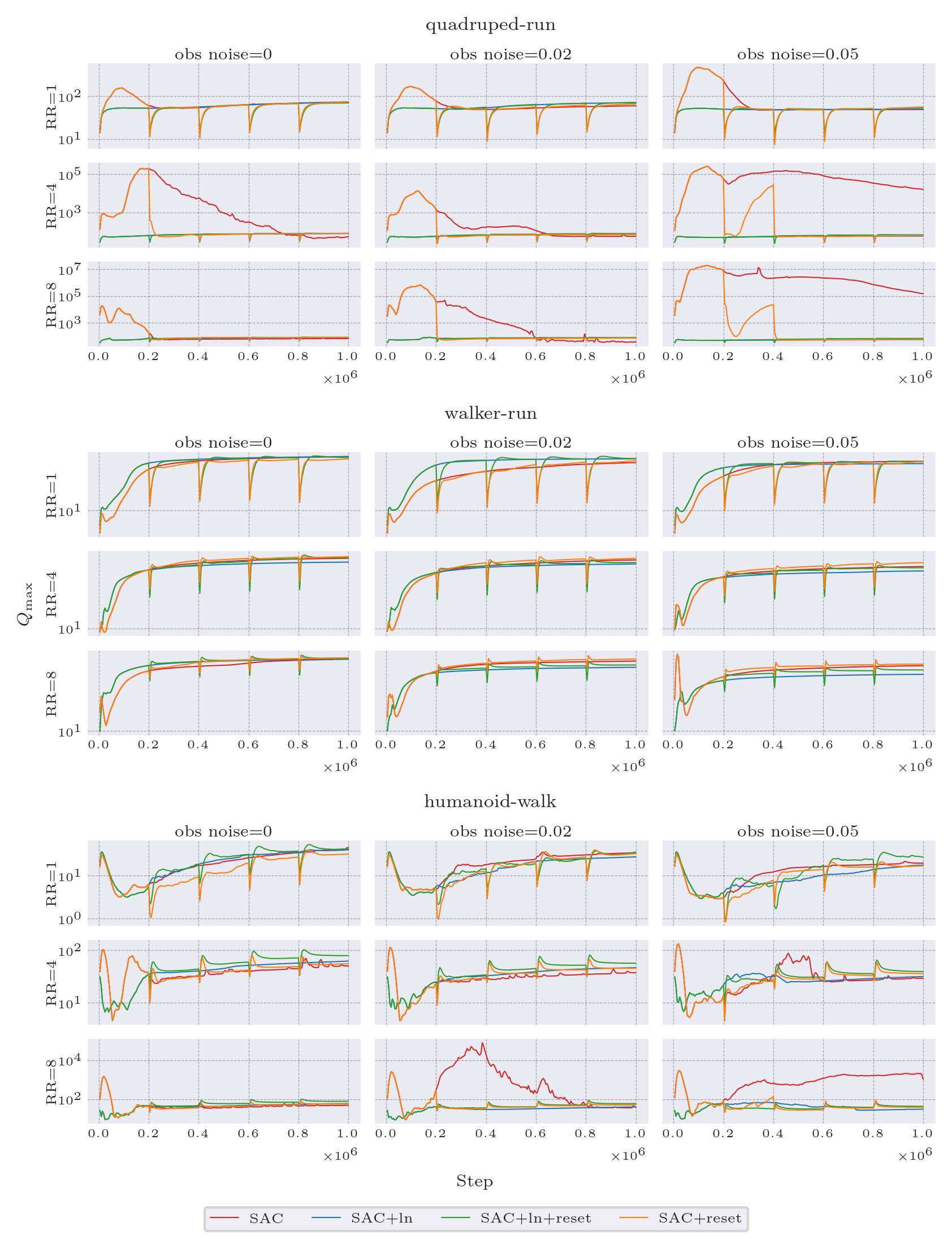}\caption{Learning curves for the $\mathcal{Q}_{max}$ across replay ratios and observation noise levels.}\label{fig:app_mq_lc_on}
\end{figure}

\begin{figure}[hb!]
  \centering
  \includegraphics[width=\textwidth]{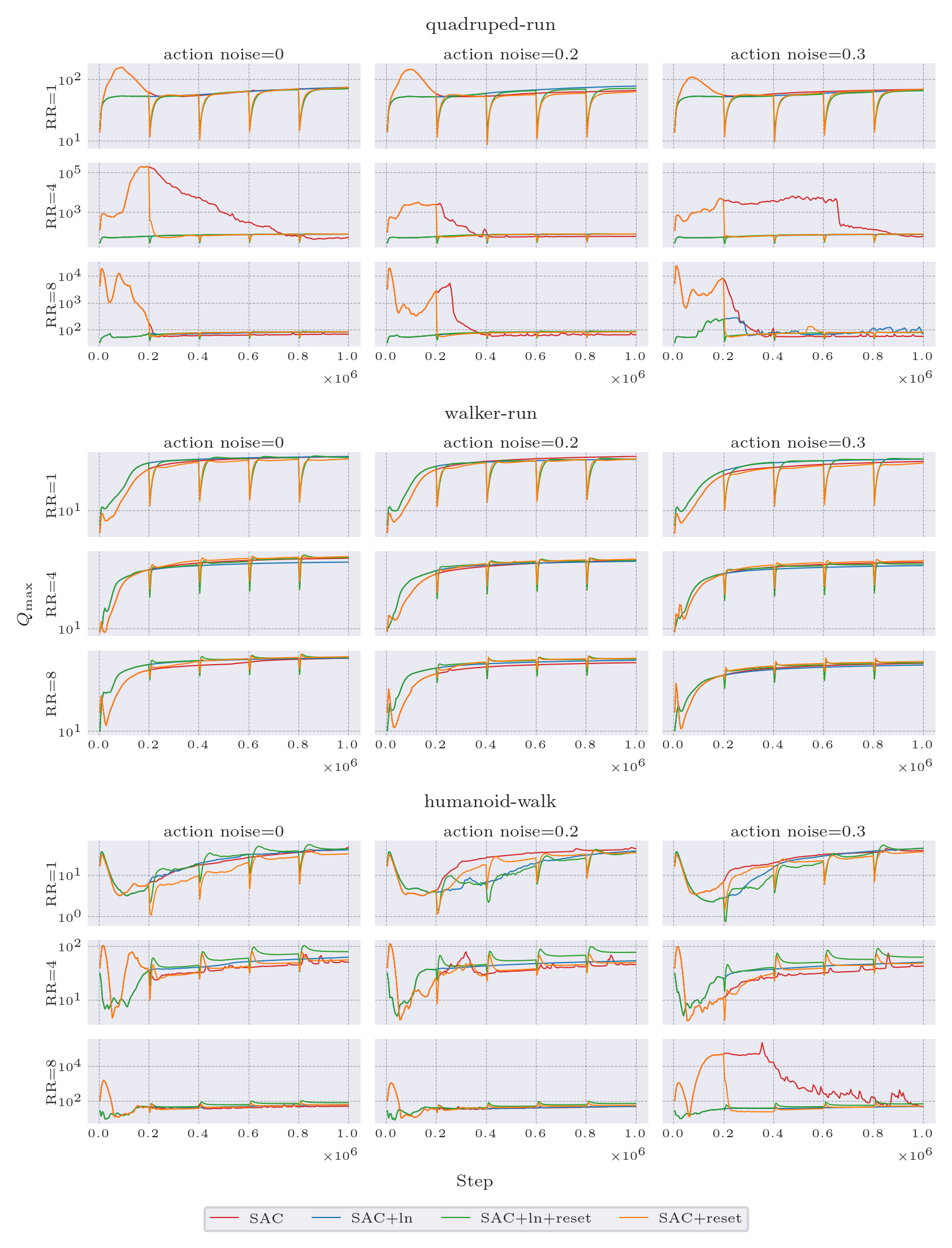}\caption{Learning curves for the $\mathcal{Q}_{max}$ across replay ratios and action noise levels.}\label{fig:app_mq_lc_an}
\end{figure}

We first provide the reward curves per considered replay ratio, environment and observation noise (\Cref{fig:app_rr_lc_on}) as well as action noise (\Cref{fig:app_rr_lc_an}). We further provide the plasticity (\Cref{fig:app_pe_lc_on,fig:app_pe_lc_an}) and $\mathcal{Q}_{max}$ evolution (\Cref{fig:app_mq_lc_on,fig:app_mq_lc_an}) plots.

\paragraph{Reward curves.}
\Cref{fig:app_rr_lc_on} shows the reward curves across observation noise levels.
\clearpage
\Cref{fig:app_rr_lc_an} shows the reward curves across action noise levels.
\clearpage
\paragraph{Plasticity evolution.}
\Cref{fig:app_pe_lc_on} shows the evolution of critic plasticity across observation noise levels.
\clearpage
\Cref{fig:app_pe_lc_an} shows the evolution of critic plasticity across action noise levels.
\clearpage
\paragraph{Maximum $\mathcal{Q}$-value.}
\Cref{fig:app_mq_lc_on} shows the evolution of $\mathcal{Q}_{max}$ across observation noise levels.
\clearpage
\Cref{fig:app_mq_lc_an} shows the evolution of $\mathcal{Q}_{max}$ across action noise levels.
\clearpage
\paragraph{Normalized scores.}
Here we report the final episode returns for the best replay ratio across all prior combinations.
\begin{figure}[hb!]
  \centering
  \includegraphics[width=\textwidth]{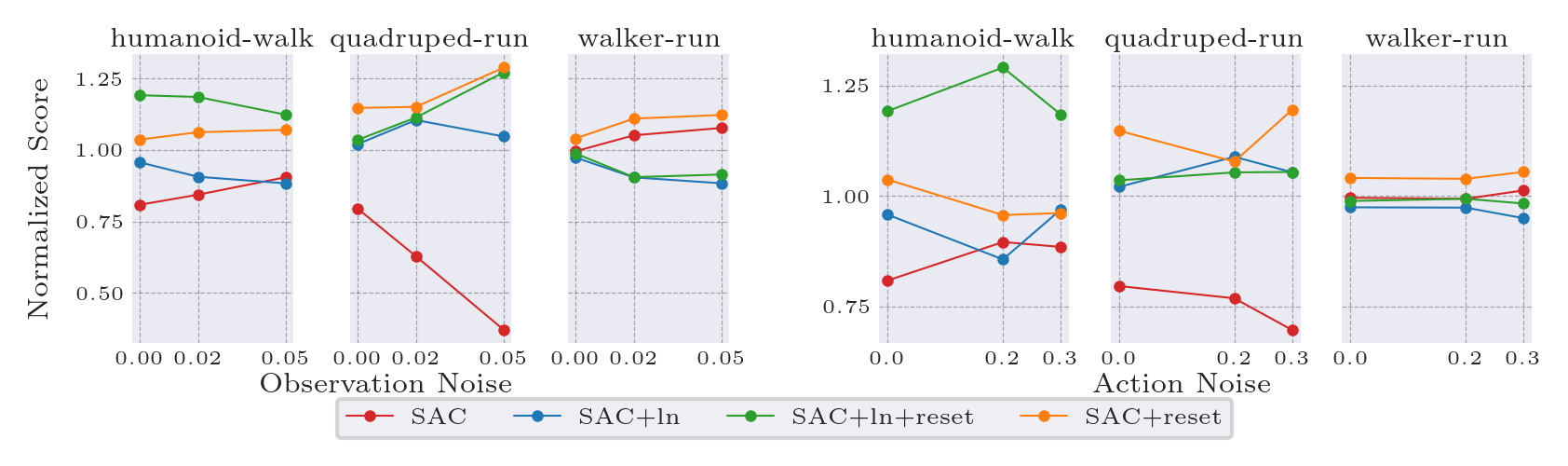}
  \caption{Normalized final episode return of the best replay ratio per method, environment and noise level.}\label{fig:normalized_best_per_method_per_env}
\end{figure}

\clearpage
\section{Noise Effect on Transition Dynamics}\label{app:noise_on_dynamics}
\begin{figure}[h!]
  \centering
  \includegraphics[width=\textwidth]{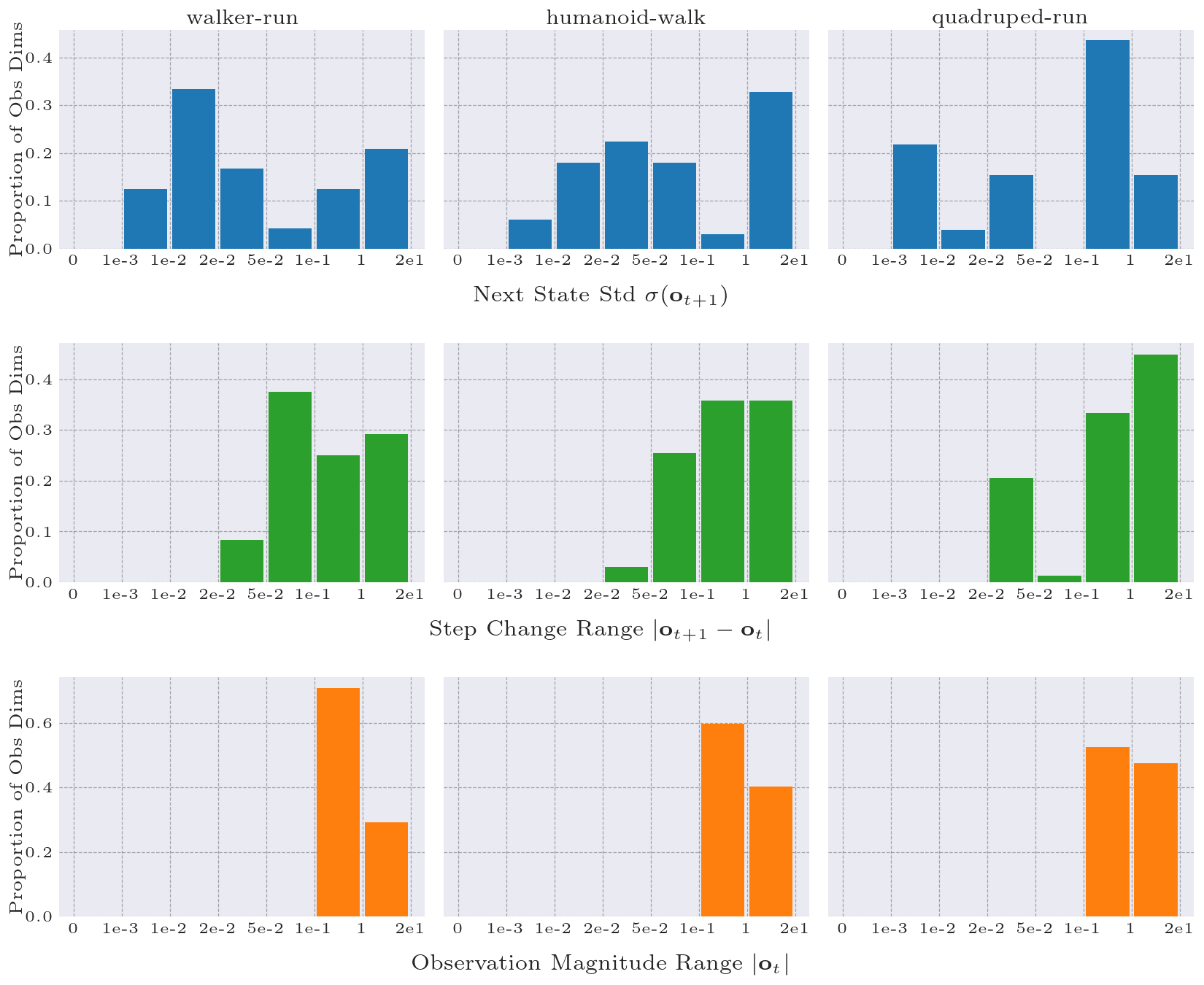}
  \caption{
  Distribution of per-dimension next-state observation sensitivity to action noise for three \texttt{dm\_control} environments.                                          
  For each step along a trajectory, $n=10$ independent action--noise pairs $(\mathbf{a}, \mathbf{a}+\boldsymbol{\varepsilon})$,                                          
  $\boldsymbol{\varepsilon}\sim\mathcal{N}(\mathbf{0},\sigma^2 I)$ with $\sigma=0.3$, are evaluated from the same physics state.                                         
  The per-step standard deviation of the resulting observation errors $f(s,\mathbf{a}+\boldsymbol{\varepsilon})-f(s,\mathbf{a})$ is computed for each observation        
  dimension and averaged over steps (1000 steps $\times$ 3 seeds).                                                                                                       
  Bars show the proportion of observation dimensions whose mean std falls within each magnitude bin.
  }
\end{figure}

\clearpage
\section{Plasticity vs.\ Learning Robustness Under Action Noise}\label{app:plasticity_vs_robustness_anoise}
\begin{figure}[h!]
  \centering
  \includegraphics[width=\textwidth]{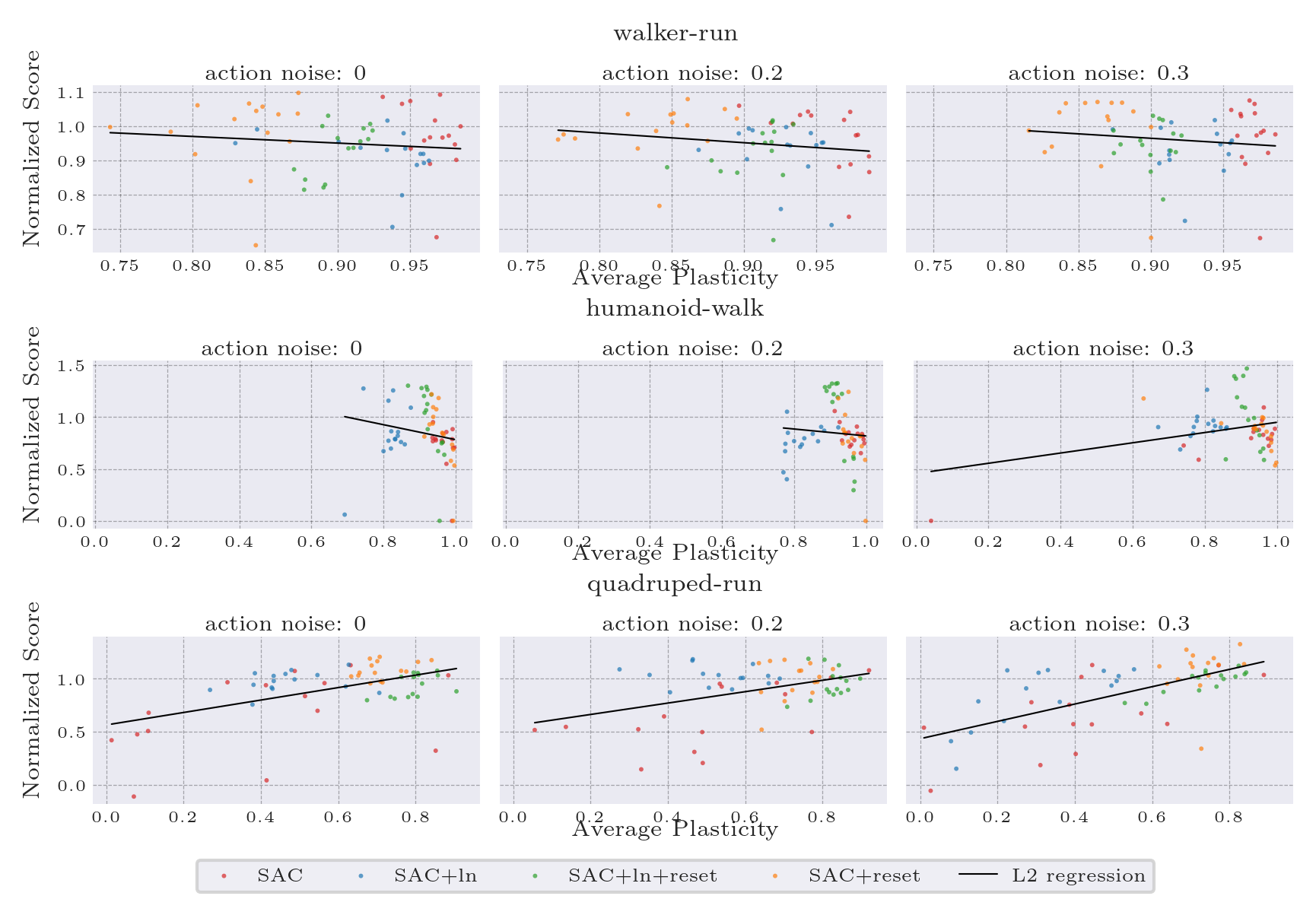}
  \caption{Mean critic plasticity over training vs.\ final normalized score, per environment and action noise level ($\sigma_\text{a}$). Within one plot each point is a combination of (method, seed, replay ratio). A $L2$-regression line is fit across all runs per plot.}\label{fig:plasticity_vs_performance_anoise}
\end{figure}

\clearpage
\section{Replay-Ratio Sensitivity to Environment Noise}
\begin{figure}[h!]
  \centering
  \includegraphics[width=\textwidth]{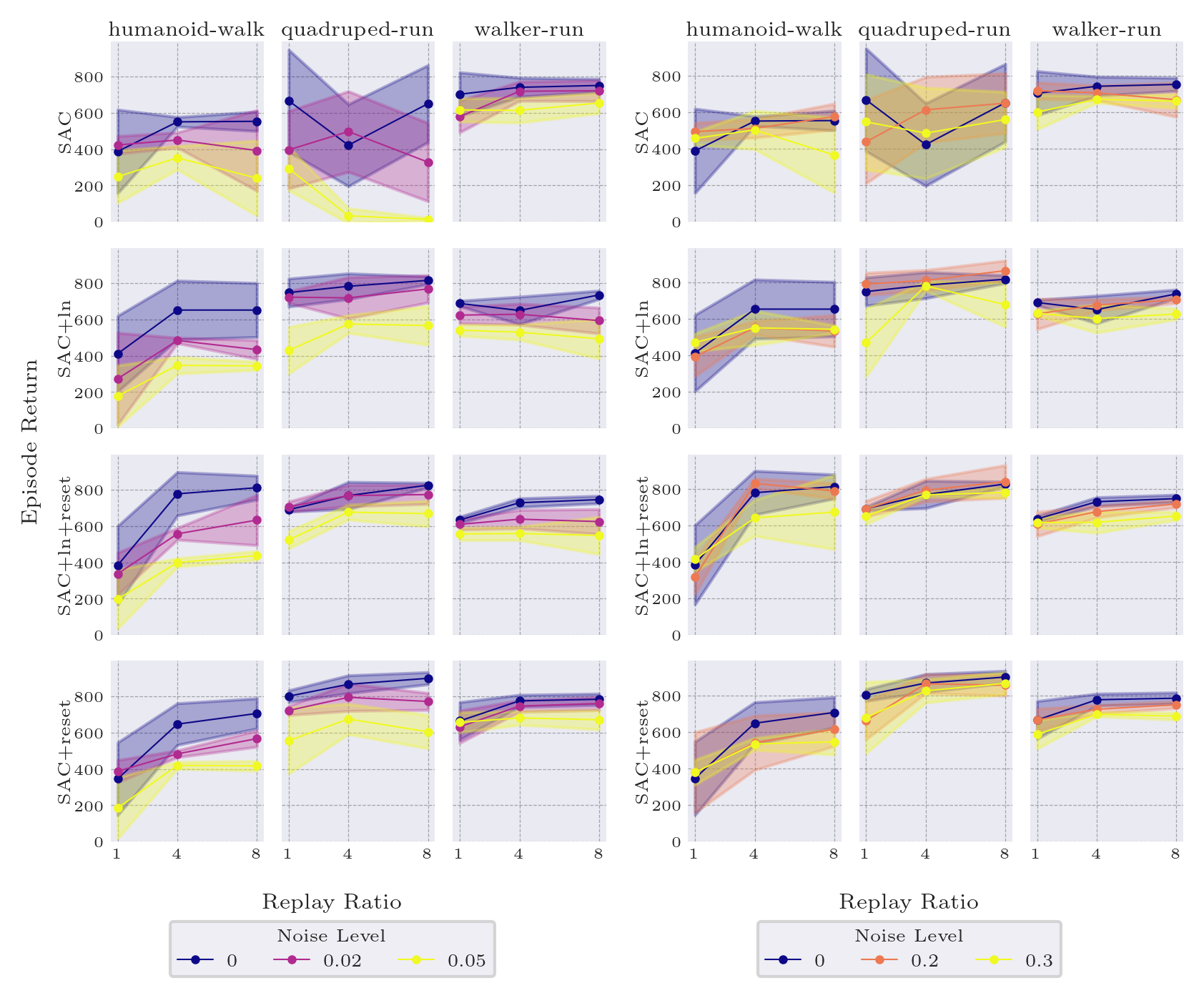}
  \caption{Episode returns from final evaluation after $10^6$ steps per noise level and across replay ratios. The goal is to detect whether high replay ratios are more affected by high noise levels due to overfitting (which is generally not the case). Each point aggregates to mean and std over 5 seeds.}
\end{figure}

\end{document}